\documentclass{article}
\usepackage{ijcai22}

\usepackage{times}
\usepackage{soul}
\usepackage{url}
\usepackage[hidelinks]{hyperref}
\usepackage[utf8]{inputenc}
\usepackage[small]{caption}
\usepackage{graphicx}
\usepackage{amsmath}
\usepackage{amsthm}
\usepackage{booktabs}
\usepackage{algorithm}
\usepackage{algorithmic}
\usepackage[switch]{lineno}

\usepackage{multirow}
\usepackage{amsthm}
\newtheorem{definition}{Definition} 
\usepackage{wrapfig}

\usepackage{subfigure}

\usepackage{algorithm}
\usepackage{algorithmic}
\usepackage{amssymb}



\usepackage[dvipsnames]{xcolor}
\usepackage{hyperref}
\hypersetup{
  colorlinks,
  citecolor=ForestGreen,
  urlcolor=BurntOrange}

\title{Time Associated Meta Learning for Clinical Prediction}

\author{
Hao Liu$^1$
\and
Muhan Zhang$^2$\and
Zehao Dong$^1$\and
Lecheng Kong$^1$\and \\
Yixin Chen$^1$\and 
Bradley Fritz$^1$\and
Dacheng Tao$^3$\And
Christopher King$^1$
\affiliations
$^1$Washington University in St. Louis\\
$^2$Peking University\\
$^3$JD Explore Academy
\emails
\{liuhao, zehao.dong, jerry.kong, ychen25, bafritz, christopherking\}@wustl.edu,\\
muhan@pku.edu.cn,
dacheng.tao@gmail.com
}

\begin{document}

\maketitle

\begin{abstract}

Rich Electronic Health Records (EHR), have created opportunities to improve clinical processes using machine learning methods. Prediction of the same patient events at different time horizons can have very different applications and interpretations; however, limited number of events in each potential time window hurts the effectiveness of conventional machine learning algorithms. We propose a novel time associated meta learning (TAML) method to make effective predictions at multiple future time points. We view time-associated disease prediction as classification tasks at multiple time points. Such closely-related classification tasks are an excellent candidate for model-based meta learning. To address the sparsity problem after task splitting, TAML employs a temporal information sharing strategy to augment the number of positive samples and include the prediction of related phenotypes or events in the meta-training phase. We demonstrate the effectiveness of TAML on multiple clinical datasets, where it consistently outperforms a range of strong baselines. We also develop a \textbf{MetaEHR} package for implementing both time-associated and time-independent few-shot prediction on EHR data.

\end{abstract}

\section{Introduction}

The use of rich Electronic Health Record (EHR) data has led to an explosion of clinical prediction for a wide variety of applications~\cite{8086133}. EHR data contains extremely detailed information about patients, such as medical history, vital signs, surgical events, and demographic information~\cite{publhealth}. Machine learning (ML) models have been proposed for many clinical outcomes, such as mortality~\cite{lagu2016validation}, chronic disease~\cite{liu2018deep}, heart failure~\cite{panahiazar2015using,8245772}, and readmission~\cite{rojas2018predicting}.             

Because of certain characteristics of EHRs including heterogeneity and noise~\cite{publhealth}, the use of EHR data in healthcare predictive models can present some unique challenges. One challenging aspect of EHR data is the limited number of available annotated examples. Even though a huge number of records may be available, many important adverse events are uncommon, and as a result, training of ML models such as neural networks can be fragile and prone to overfitting. Label sparsity is aggravated by the many complex and high-dimensional features present in EHRs. 

In this paper, we address a very common few-shot problem in clinical prediction: \textbf{prediction of future events in specific time-ranges (time-associated target prediction)}. Specifically, a clinical event (such as deterioration or improvement) can occur in different time periods, and the likely lead-time until these events greatly affects the interpretation and is vital for users.
For example, predicted clinical deterioration could trigger immediate changes in therapy or consumption of scarce resources (such as ICU space) or watchful waiting and more tests depending on if the event is likely in the near or far term.
In our example of postoperative survival time prediction, death soon after surgery has very different management implications from death in the subsequent weeks or months.
Other events (such as discharge from hospital) are inevitable, and the appropriate planning and patient preparation steps depend on the likely lead-time.

Time-associated target prediction can be approached using explicit time-to-event models~\cite{wang_mortality_2015}.
However, directly learning to predict the time to an event is limited by (1) many patient records being censored by loss to follow up before the event, (2) a limited ability to tune the algorithm to focus on ``more relevant'' time windows or clinically equivalent time windows, (3) parametric assumptions on the distribution of event times built into loss functions, (4) difficulty interpreting the expected event time in the context of multi-modal outcome distributions. 
Others have used multi-task learning to predict events at different times as joint tasks~\cite{li_deepalerts_2020,multi1}.
However, traditional multi-task learning~\cite{multi1} requires identical low-level representations of data across tasks, which may be too strong an assumption.

To bridge the gap in learning ability between people and AI for small samples, a new paradigm was proposed named few-shot learning~\cite{few-shot1}. Meta-Learning has been shown to be an effective method to deal with few-shot problems, most of which rely on learning initialized parameters~\cite{MAML} and a metric space~\cite{snell2017prototypical} across tasks. 
Meta learning assumes that an internal data representation is transferable between tasks~\cite{suo2020tadanet}, and as discussed in \cite{jose2021information},  existing meta learning models tend to work better when meta-training tasks are similar to the target. However, in predicting the occurrence time of a clinical event under a regression framework, few tasks with similar distributions are usually available to directly apply meta learning.

To address the above challenges, we propose \textbf{Time Associated Meta Learning (TAML)}, a novel adaptation of model-agnostic meta-learning (MAML) ~\cite{MAML} for temporal problems. We assume that the ultimate goal is to predict the occurrence of a target event in a specific time period. We split a time-predictive problem into multiple binary classification problems corresponding to events in different time periods, thus generating many similar tasks. 
To the best of our knowledge, this is the first study to transform the time-associated few-shot problem into a meta learning problem where we not only split the problem into multiple similar tasks, which are called \textbf{time-associated tasks}, but also include more time-independent tasks as meta learning tasks, which are called \textbf{reference tasks}. Considering the effect of task similarity on prediction performance, we distinguish these two kinds of tasks in the training process and apply different weighting strategies to emphasize the importance of time-associated tasks. This allows reference tasks to provide directions during training without overwhelming the time-associated tasks. 
Our approach keeps model weights for each task within a tunable distance of each other, encouraging joint learning but allowing differences, conceptually similar to fine-tuning each task during training, and weakening the necessary assumptions compared to multi-task learning.

Such a setup results in fewer positive samples for each classification task. To overcome the drawback of sparsity and to incorporate inter-task timing information into this classification problem, we develop a novel task partitioning strategy named the \textbf{T}emporal \textbf{I}nformation \textbf{S}haring \textbf{S}trategy (TISS) to augment the positive samples by exploiting the temporal persistence of the patient-state. 
We propose four situations for a patient-state in one time period, where we label the existence of patient-state in the current time period as a positive situation during training, thus augmenting positive labels. TISS is also refined to the different treatment of tasks belonging to different situations with the same label. 
            
Our main contributions are: 1) We are the first study to transform time-associated regression problems to a MAML setting with joint time-independent tasks. 2) We propose TISS to exploit temporal dependence of health states and augment positive labels in training. 3) We provide a general TAML framework with sufficient details to allow any time-associated clinical prediction task to benefit from it. We show results on two public datasets and a real application on a local dataset. We use two clinical events as labels: mortality and intensive care unit discharge, where TAML outperforms state-of-art baselines.
The TAML approach turns out to be remarkably insensitive to tuning parameters that are critical in other approaches; dividing time into very fine windows is harmless, and adding tasks of unclear relevance to the ensemble has very little risk.
4) We provide a package for few-shot clinical prediction named MetaEHR, which includes both the implementation of TAML for time-associated targets and other meta learning algorithms for time-independent targets.

\section{Related Works}

In contrast to traditional supervised machine learning algorithms, which learn a model for each label, meta-learning is a strategy of learning to learn.
The goal is to train a model on a variety of tasks and use this experience to improve future learning performance. Because of the strong inductive bias created, a meta-trained algorithm can be applied to solve few-shot problems. 

Metric-based meta learning and optimization-based meta learning are the two main categories of meta learning.
Metric-based meta learning aims to learn a metric or distance that compares training data with testing data. For example, \cite{koch2015siamese} proposed a method to use a Siamese Neural Network for one-shot image classification. 
The Siamese Neural Network is composed of two twin networks, the outputs of which are combined with a function for learning the relationship between pairs of input data examples. 
Prototypical Networks~\cite{snell2017prototypical} encode each input into a continuous latent space and carry out classification using the similarity of an example to the representation of latent classes. 

Optimization-based meta learning is essentially learning a good initialization of a neural network from which fine-tuning on a small number of additional training examples can be effective. 
MAML~\cite{MAML} is a widely used optimization-based meta-learning technique. Briefly, MAML trains in a nested loop.
At each outer loop iteration, a set of tasks is sampled.
In the inner loop, the current set of global parameters is updated with one or a few gradient steps \textit{independently for each task}.
The global parameters are then updated using a loss function which sums over all tasks \textit{including the adaptation computed in the inner loop}.
The global parameters therefore evolve to a point from which an acceptable model for any of the training tasks can be reached with a small step.
This is made feasible by similar low-level representations and latent structures between tasks, which is also an assumption of the method.
Since this method can be applied to diverse classes of models, it has been used in image classification~\cite{raghu2019rapid}, reinforcement learning~\cite{liu2019taming} and a variety of other domains. 

Multi-task learning addresses a similar need.
By simultaneously training a base model and multiple ``heads'' corresponding to outcomes, common labels allow an algorithm to learn an effective data-representation and a relatively low-complexity head for rare outcomes.
For example, \cite{liu2020multi} uses MTL to predict mortality of rare diseases. 
Other meta-learning approaches include MetaPred~\cite{zhang2019metapred} which uses a simulated target during the meta-train process, and MetaCare++~\cite{tan2022metacare++} which proposes a specialized clinical meta-learner with a hierarchical subtyping strategy to capture temporal relations.

\section{Time Associated Meta Learning (TAML)}
In this section, we describe the TAML framework. We introduce the meta-learning problem setup and the meta-train and meta-test phases, and explain the Temporal Information Sharing Strategy (TISS).

\subsection{Meta-Learning Problem Setup} 
The purpose of TAML is to predict the occurrence of patient events during potentially multiple time periods of interest. We use survival time after surgery as an example. Since death shortly after most kinds of surgery is uncommon, it is hard to train an accurate prediction model, and our goal is to use adjacent time periods to improve model performance.

However, viewing this few-shot problem from a regression perspective, i.e., calculating the expected event time, would leave the meta learning without enough relevant tasks. A limited number of patient events are time-associated and fully recorded, and MAML benefits from more closely related training tasks ~\cite{jose2021information}.

\begin{figure}
    \centering
    \includegraphics[width=0.5\textwidth]{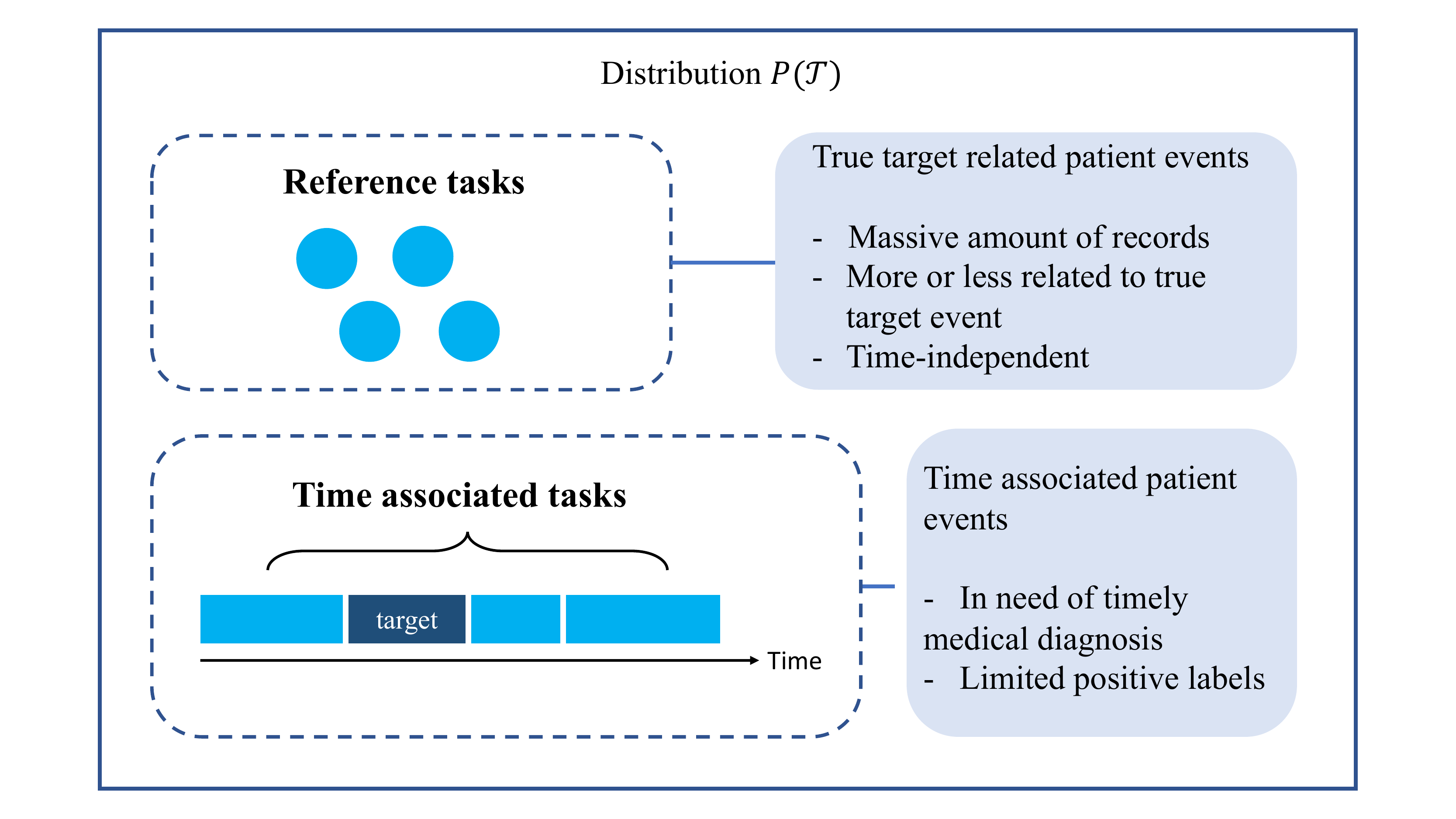}
    \caption{Task Decomposition. Reference tasks are on the top. They are time-independent and related to the true target. Time-associated tasks are at the bottom. Each target is represented by a block over the timeline.}
    \label{fig:tasks}
    \vspace{-10pt}
\end{figure}

We therefore generate new classification tasks by grouping patient events according to time of occurrence, where each task is a binary prediction of the occurrence of that event during a certain time period. This way, we effectively increase the number of available tasks. These \textbf{time-associated tasks} are highly related to each other, which is the setting in which meta learning has been successful. Additionally, since the problem has been switched into a binary classification form, other time-independent diseases or events related to the target with a larger number of positive labels can be added to the task space. We will refer to these additional tasks as \textbf{reference tasks}. The available tasks are depicted in Figure \ref{fig:tasks}.

In a fashion similar to MAML~\cite{finn2017meta}, in TAML we train a meta model which can quickly adapt to tasks that follow some distribution $p(\mathcal{T})$. The model is denoted by $f$ with parameters $\theta$, and the model $f$ maps an input data $\mathbf{x}$ into a binary value $y$. There are two kinds of tasks involved in the TAML framework, time-associated tasks $\{\mathcal{T}^S\}$ and reference tasks $\{\mathcal{T}^R\}$. Both groups of tasks will be used in our framework, but the time-associated tasks are what we really want to predict. The time-associated tasks, reference tasks, meta-train phase, and meta-test phase that will be used in our framework are defined below.

\begin{definition}
\label{def:3.1}
$\mathcal{T}_0$ is the \textbf{true target} patient event. 
We assume each example of $\mathcal{T}_0$ is associated with a timestamp. We divide the given time span into $J$ non-overlapping time periods.
$\{\mathcal{T}^S_j\}$ ($j=1,2,\ldots, J$) is the set of \textbf{time-associated tasks}, which includes the occurrence of $\mathcal{T}_0$ in different time periods. For example, $\mathcal{T}^S_j$ describes whether $\mathcal{T}_0$ occurs in the $j^{th}$ time period. $\{\mathcal{T}^R_i\}$ denotes the set of \textbf{reference tasks}, which includes other patient events potentially related to $\mathcal{T}_0$. We assume reference tasks do not contain timestamps.
\end{definition}

The framework consists of meta-train and meta-test phases. In the \textbf{meta-train} phase, parameters $\theta$ of a global-task model are trained with time-associated tasks and reference tasks. In the \textbf{meta-test} phase, we select one of the time-associated tasks as a target and fine-tune the meta-learner, evaluating the performance on a held-out test set.

\subsection{The Proposed Framework} 
TAML training is very similar to MAML~\cite{MAML}. 
The parameters of the meta-model $\theta^*$ are defined as
\begin{equation}
    \theta^{*} = \mathop{argmin}_{\theta} \mathbb{E}_{\mathcal{T}_k\sim p(\mathcal{T})}L_k(D_k',\theta_k'(\theta)),
\end{equation} 
where tasks $\mathcal{T}_k$ follow a distribution $p(\mathcal{T})$, and consist of both time-associated tasks $\mathcal{T}^S_j$ and reference tasks $\mathcal{T}^R_i$. 
$D_k'$ represents data points sampled from task $\mathcal{T}_k$. $L_k$ is a loss function for task $\mathcal{T}_k$ over data points $D_k$ and parameters $\theta_k$, which may take various forms depending on the type of problem we focus on. $\theta_k'$ is the parameters adapted to task $\mathcal{T}_k$ in the inner-level loop using initialized parameters $\theta$. The intuition is that we aim to find the meta parameters $\theta^{*}$ that have a small loss on every task $\mathcal{T}_k$ after adapting to that task. This requires $\theta^{*}$ to absorb information from all the tasks and be adaptable to them.
Algorithm 1 includes the architecture of the meta-train phase of TAML framework, which is composed of inner-level and outer-level loop updates.

\begin{algorithm}[t]
\caption{Time-associated meta-train}
\textbf{Input}: Time-associated tasks $\{\mathcal{T}^S\}$, Reference tasks $\{\mathcal{T}^R\}$ \\
\textbf{Parameter}: Inner step sizes $\alpha_k$, outer step size $\beta$, outer task weights $w_k$\\
\begin{algorithmic}[1]
\STATE Randomly initialize $\theta$
\WHILE{Outer Loop not done}
\STATE Sample batch tasks $\{\mathcal{T}_k\}$ including $\{\mathcal{T}_k^S\}$ and $\{\mathcal{T}_k^R\}$
\FORALL {$\{\mathcal{T}_k\}$}
\STATE Sample data points $D_k$ and $D_k'$ from $\mathcal{T}_k$ 
\STATE Compute Loss function with $D_k$ and $\theta$
\STATE Use gradient descent to calculate $\theta_k'$ with $\alpha_k$ 
\STATE $\theta'_k = \theta - \alpha_k\bigtriangledown_\theta L(D_k,\theta)$
\ENDFOR
\STATE Calculate weighted loss functions $L_k$ for each task $\{\mathcal{T}_k\}$ using $D_k'$ and $\theta_k'$
\STATE $L_k(D_k',\theta_k') = w_kL(D_k',\theta_k')$
\STATE Update $\theta = \theta - \beta\bigtriangledown_\theta \sum_kL_k(D_k',\theta_k')$

\ENDWHILE
\end{algorithmic}
\end{algorithm}

In each iteration of the outer-level loop, given a $\theta$ which is randomly initialized at the first time and later provided by the last iteration, we first uniformly sample a batch of tasks $\{\mathcal{T}_k\}$ including both time-associated tasks $\{\mathcal{T}_k^S\}$ and reference tasks $\{\mathcal{T}_k^R\}$ from the task distribution $p(\mathcal{T})$. 

Then, in the inner-level loop, for each task, we sample two sets of data points, $D_k$ and $D_k'$, which form the \textbf{support set} and \textbf{query set}. $D_k$ is used in the inner-level loop update to obtain the adapted parameters $\theta_k'$, where TISS labeling method is applied to $D_k$. $D_k'$ is used in the outer-level loop update to compute overall training objective for the update of global parameters $\theta$. Formally, the inner-level loop update is given by,
\begin{equation}
\theta'_k = \theta - \alpha_k \nabla_\theta L(D_k,\theta),   
\end{equation}
where $\alpha_k$ is a task-specific update step size. The value of $\alpha_k$ depends on the task category. After the inner loop, $\theta_k'$ is obtained for each task $\mathcal{T}_k$, which simulates a fine-tuning process to adapt the meta-learned $\theta$ to each specific task.

Since the reference tasks and the time-associated tasks have different relationships to the true target, we give them different step sizes $\alpha_k$ to reflect their relative importance. Considering the true target is time-associated, parameters of each time-associated task should provide more information than reference tasks, thus larger step size $\alpha_k$ is assigned to time-associated tasks.

The outer-level update optimize $\theta$ with the sum of weighted gradients on $D_k'$ with respect to $\theta_k'$. Similarly, we place a greater emphasis on the loss functions on k in $\mathcal{T}^S$ than $\mathcal{T}^R$ when updating $\theta$. We use the following equation to calculate a loss function that is advantageous for updating parameters $\theta$ to a direction closer to that of time-associated tasks.
\begin{equation}\label{main_update}
    L_k(D_k',\theta_k') = w_kL(D_k',\theta_k').
\end{equation}  
The selected weight $w_k$ is a hyper-parameter determined by both the task type and the time period. Specifically, $w_k$ is constant for all tasks in $\{\mathcal{T}^R\}$, but different (and larger) for tasks in $\{\mathcal{T}^S\}$, similar to when selecting the hyper-parameters $\alpha_k$.
Then the sum of loss functions $L_k$ will be used to update the parameter $\theta$ by equation \ref{theta_update}.
\begin{equation}\label{theta_update}
    \theta = \theta - \beta \nabla_{\theta} \sum_k L_k(D_k',\theta_k'),
\end{equation}
where $\beta$ is the outer-level loop learning rate. 
Note that the derivative with respect to $\theta$ propagates through $\theta_k'$, requiring an additional back-propagation through the computation graph for calculating $\theta_k'$, which is supported by standard deep learning libraries.

 After obtaining a good initialization of parameters from the meta-train phase, the goal of the meta-test phase is to fine-tune a model that can predict the occurrence of the true target patient event in a specific time period. 
 In the meta-test phase, the target event that occur in a certain time period is collected to form the test task. We adapt the initialized parameters learned from the meta-train phase to the test task.

\subsection{Temporal Information Sharing Strategy}

We previously described how to transform time-associated regression predictions into classification predictions suitable for MAML. In this new setting, a few time-associated tasks and reference tasks are available for training. However, these time-associated tasks essentially deal with a classification problem with extremely limited positive labels due to the time-based division strategy in Def \ref{def:3.1}. Hence, the problem inherently poses the challenge of label imbalance.

\begin{figure}
    \centering
    \includegraphics[width=0.49\textwidth]{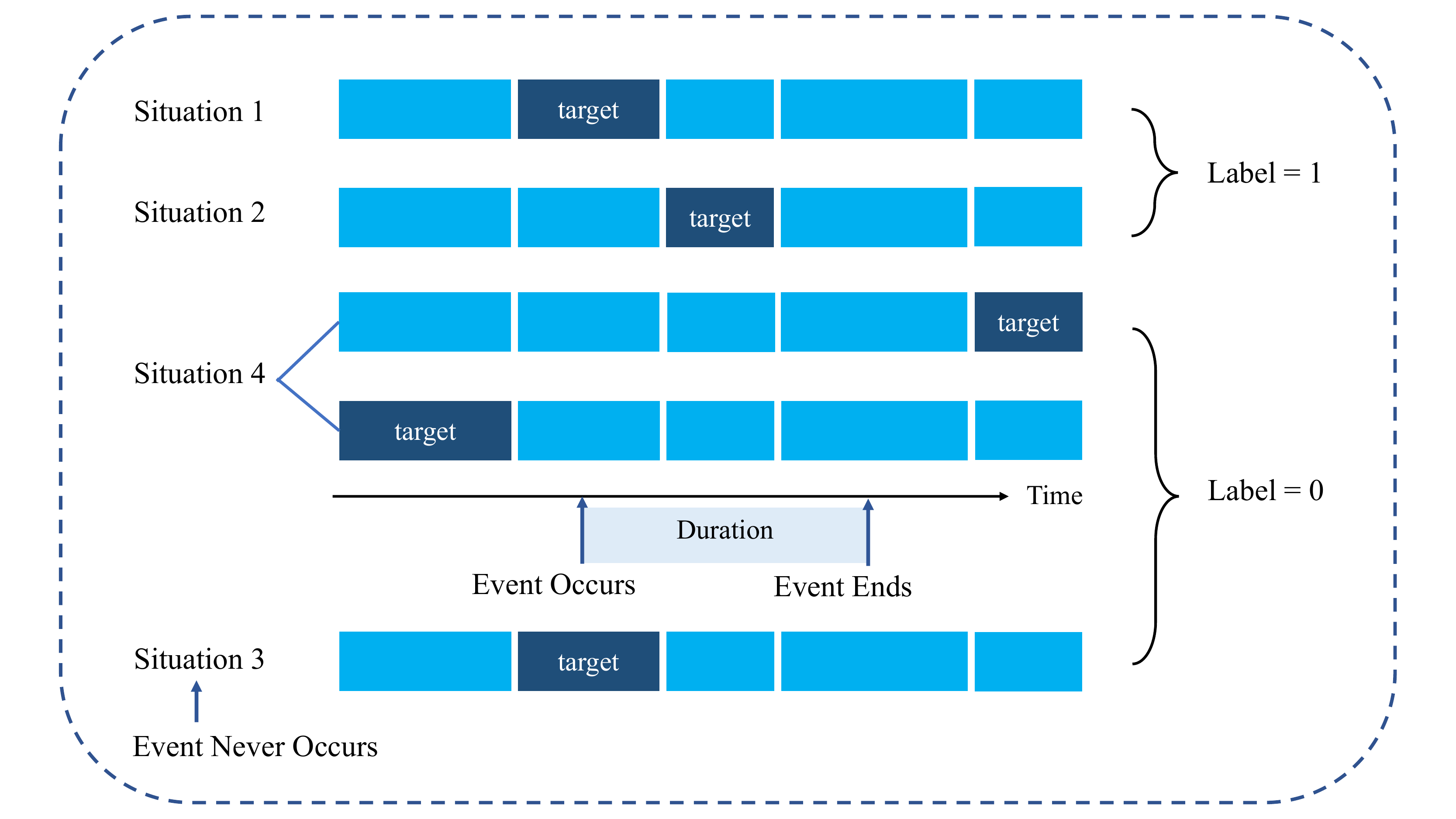}
    \caption{TISS Strategy. In the top three rows, the event  occurrence is in the second time period and lasts until the fourth time period. The dark blue block represents the target time period. Bottom row: no events occur. }
    \label{fig:algorithm}
    \vspace{-10pt}
\end{figure}

To address the positive label sparsity issue, we propose a strategy to augment the number of positive labels, \textbf{T}emporal \textbf{I}nformation \textbf{S}haring \textbf{S}trategy (TISS), which incorporates the temporal persistence of the patient event, thus taking the impact of other time periods on the target time period into consideration. Basically, TISS is a situation-based label setting strategy for time-associated tasks $\{\mathcal{T}^S_j\} ($j=1,2,\ldots, J$)$. There are in total \textbf{four situations} discussed, and Figure \ref{fig:algorithm} illustrates these situations with corresponding binary labels.

Among the four situations, $S_1$ represents that the patient event precisely occurs during the target time period. $S_2$ indicates that the patient event occurred prior to the target time period and lasted at least until the target time period. $S_3$ means the patient event never occurs during the entire observation. $S_4$ which has two possible sub-situations, denotes the situation that the patient event begins after the target time period or ends before the target time period.

\begin{table*}[t]
\small
\centering
\begin{tabular}{lcccccccc}

\toprule
\multicolumn{1}{c}{\multirow{2}{*}{Model}} & \multicolumn{2}{c}{0-6 Days} & \multicolumn{2}{c}{7-18 Days} & \multicolumn{2}{c}{19-30 Days} & \multicolumn{2}{c}{31-90 Days} \\\cmidrule(l){2-3}\cmidrule(l){4-5}\cmidrule(l){6-7}\cmidrule(l){8-9}
\multicolumn{1}{c}{}                       & AUROC         & Recall       & AUROC         & Recall        & AUROC          & Recall        & AUROC          & Recall        \\
\midrule
DNN                                        & 0.7590        & 0.4318       & 0.7523        & 0.4463        & 0.7476         & 0.4379        & 0.7653         & 0.4662        \\
Multi-task Learning                        & 0.7783        & 0.4598       & 0.7892        & 0.4832        & 0.7912         & 0.4965        & 0.7734         & 0.4475        \\
Pre-trained                                & 0.7684        & 0.4381       & 0.7796        & 0.4715        & 0.7834         & 0.4869        & 0.7751         & 0.4637        \\
Survival Analysis
& 0.7590    & 0.4369    & 0.7642  & 0.4614    & 0.7785  & 0.4781    & 0.7648  & 0.4703
\\

Prototypical Network
& 0.7746    & 0.4391    & 0.7813    & 0.4698    & 0.7861   & 0.4863    & 0.7775    & 0.4714
\\

MAML                                       & 0.7689        & 0.4487       & 0.7732        & 0.4786        & 0.7852         & 0.4881        & 0.7813         & 0.4763        \\
\textbf{TAML(ours)}                                 & \textbf{0.8093} & \textbf{0.4713} & \textbf{0.8179} & \textbf{0.4987} & \textbf{0.8228} & \textbf{0.5007} & \textbf{0.8137} & \textbf{0.4963}\\       
\bottomrule
\end{tabular}
\caption{Performance Evaluation of Survival Time on \textit{local hospital} dataset}
\label{table1}
\vspace{-5pt}
\end{table*}
\begin{table}[t]
\centering
\small
\begin{tabular}{lcccccc}
\toprule
\multicolumn{1}{c}{\multirow{2}{*}{Model}} & \multicolumn{2}{c}{0-6 Hours}     & \multicolumn{2}{c}{6-12 Hours}    & \multicolumn{2}{c}{12-24 Hours}   \\\cmidrule(l){2-3}\cmidrule(l){4-5}\cmidrule(l){6-7}
\multicolumn{1}{c}{}                       & AUC           & Recall          & AUC           & Recall          & AUC           & Recall          \\
\midrule
DNN                                        & .7867          & .4254          & .7633          & .4179          & .7809          & .4325          \\
MTL                        & .8371          & .4654          & .8318          & .4679          & .8359          & .4730           \\
Pre-train                                & .8032          & .4568          & .7963          & .4493          & .8137          & .4634          \\
Survival
& .7892    & .4417  & .7651  & .4308  & .7813 & .4305
\\
ProtoNet
& .8346  & .4579  & .8303  & .4691  & .8410  & .4724
\\
MAML                                       & .8297          & .4598          & .8216          & .4571          & .8304          & .4694          \\
\textbf{TAML}                        & \textbf{.8643} & \textbf{.4891} & \textbf{.8572} & \textbf{.4870} & \textbf{.8697} & \textbf{.4958}\\
\bottomrule
\end{tabular}
\caption{Performance of Survival Time on MIMIC III}
\label{table2}
\vspace{-5pt}
\end{table}

Furthermore, we divide the four situations into two categories: the first category is referred to as the \textbf{absolute real situations}, including $S_1$ and $S_3$; the second category is referred to as the \textbf{auxiliary situations}, including $S_2$ and $S_4$. 
In TISS, in addition to the label flipping of $S_2$, we emphasize the importance of absolute real situations by assigning higher weights to the examples of absolute true situations. We refer to the weight ratio of the first category of situations over the second category as the \textbf{augmentation ratio}, where a larger augmentation ratio means the algorithm focuses more on the examples of absolute real situations, and vice versa.

Some patient events (like ICU admission) have real associated durations.
However, events like death never ``resolve'', and other patient states have unclear or unrecorded actual durations.
We can adopt the same computational strategy with these kinds of outcomes by assigning them a pseudo-duration; effectively, smoothing events from one time period into adjacent ones. 

For $S_2$, while giving it a positive label, we limit its ability to influence the model. For $S_4$, although we retain its negative label, the fact that the patient event has occurred in other time periods also reflects the patient's health status, so the two negative label situations, $S_3$ and $S_4$, are distinguished.
With durations set to a large number, the method reduces to estimating the cumulative event rate at each time.

The TISS strategy will be applied to the training  process in both the meta-train (inner-loop) and meta-test (fine-tuning) phases, and the original occurrence labeling method will be used in the outer-level loop of the meta-train phase and evaluation of the meta-test phase.


\section{Experimental Results}

We first introduce the Python package for few-shot clinical prediction named MetaEHR. Then, we conduct experiments to predict two clinical events: time to death (survival time) and the time to discharge of intensive care unit (ICU) patients (length of stay = LOS). We compare the performance to state-of-the-art baselines, conduct ablation studies, and explore sensitivity to tuning parameters.

\subsection{MetaEHR: a Python Package for Few-shot Clinical Prediction}
We develop an easy-to-use Python package: MetaEHR, for few-shot clinical classification prediction. MetaEHR provides both TAML implementation on time-associated targets and several meta learning algorithm implementations on time-independent targets. 
The implementation of meta-learning on EHR data differs from other domains in that each task comes from a separate outcome where all the outcomes share common patients. Please find it in \url{https://github.com/Haoliu-cola/Meta-EHR}.

 \subsection{Labels and Dataset Description}

\textbf{Survival time} is the first label. We present results with different goals (long and short term survival) on two datasets: surgical patients at \textit{local hospital} and the Medical Information Mart for Intensive Care III database (MIMIC III). The \textbf{length of stay of ICU patients} is the second label, where we meta-train on two public datasets: MIMIC IV and eICU and show generalization performance by meta-testing on \textit{local hospital} dataset.

\begin{table}[t]
\small
\centering
\begin{tabular}{lcccccc}
\toprule
\multirow{2}{*}{Model} & \multicolumn{2}{c}{0-1 Days}      & \multicolumn{2}{c}{1-2 Days}      & \multicolumn{2}{c}{2-4 Days}      \\ \cmidrule(l){2-3}\cmidrule(l){4-5}\cmidrule(l){6-7}
                       & AUC           & Recall          & AUC           & Recall          & AUC           & Recall          \\ \midrule
DNN                    & .7243          & .4203          & .7385          & .4169          & .7019          & .4017          \\
MTL    & .7518          & .4272          & .7578          & .4294          & .7476          & .4324          \\
Pre-train            & .7372          & .4166          & .7443          & .4127          & .7342          & .4193          \\
Survival      & .7148          & .4057          & .7296          & .4204          & .7175          & .4140           \\
ProtoNet               & .7324          & .4285          & .7537          & .4261          & .7422          & .4258          \\
MAML                   & .7458          & .4209          & .7642          & .4278          & .7486          & .4236          \\
\textbf{TAML}    & \textbf{.7892} & \textbf{.4418} & \textbf{.7946} & \textbf{.4617} & \textbf{.7715} & \textbf{.4436} \\ \bottomrule
\end{tabular}
\caption{Performance Evaluation of ICU LOS}
\label{table3}
\vspace{-5pt}
\end{table}

 The dataset from \textit{local hospital} includes patient demographics, height and weight, comorbidities, preoperative vital signs, laboratory results, and medications for all adults undergoing inpatient surgery from September 2012 to June 2018. 42,853 patients with preoperative features are included in the experiment. 
 It has been described previously in \textit{anonymized references}.
 
 MIMIC III is a publicly accessible single-center critical care ICU database from the Beth Israel Deaconess Medical Center (BIDMC)~\cite{mimic}. 
It includes information on 46,521 patients admitted to ICU from 2001 to 2012. We extract both time-dependent statistical features and discrete features.
MIMIC IV~\cite{https://doi.org/10.13026/as7t-c445} is a partially overlapping dataset with MIMIC III; it excludes early years of MIMIC III (before a transition in the EHR) and includes several years after MIMIC III closed. 
The eICU Collaborative dataset~\cite{eicu,pollard2018eicu} contains EHR abstracts for more than 200,000 patients from 2014 and 2015.

In ICU LOS experiment, a set of overlapping features are selected between MIMIC IV, eICU, and \textit{local hospital} datasets, including demographic and statistical features of vital signs. The available number of EHRs is 34,925, 73,908, and 25,591 respectively.

  \begin{table*}[t]
\centering
\small
\begin{tabular}{lcccccccc}

\toprule
\multicolumn{1}{c}{\multirow{2}{*}{Model}} & \multicolumn{2}{c}{0-6 Days} & \multicolumn{2}{c}{7-18 Days} & \multicolumn{2}{c}{19-30 Days} & \multicolumn{2}{c}{31-90 Days} \\\cmidrule(l){2-3}\cmidrule(l){4-5}\cmidrule(l){6-7}\cmidrule(l){8-9}
\multicolumn{1}{c}{}                       & AUROC         & Recall       & AUROC         & Recall        & AUROC          & Recall        & AUROC          & Recall        \\
\midrule

MAML                                       & 0.7689        & 0.4487       & 0.7732        & 0.4786        & 0.7852         & 0.4881        & 0.7813         & 0.4763        \\
TAML w/o TISS                                       & 0.7790        & 0.4502       & 0.7796        & 0.4817        & 0.7913         & 0.4942        & 0.7867         & 0.4893        \\
TAML w/o weight                                       & 0.7895        & 0.4623       & 0.7941        & 0.4884        & 0.8058         & 0.4912        & 0.7957         & 0.4932        \\
TAML w/o TISS train
& 0.7942 & 0.4679 & 0.8043 & 0.4933 & 0.8107 & 0.5014 & 0.8095 & 0.4982\\
TAML w/o unrelated
& \textbf{0.8126} & \textbf{0.4857} & 0.8163 & 0.4905 & 0.8224 & \textbf{0.5033} & 0.8096 & 0.4930\\
\textbf{TAML(ours)}                                 & 0.8093 & 0.4713 & \textbf{0.8179} & \textbf{0.4987} & \textbf{0.8228} & 0.5007 & \textbf{0.8137} & \textbf{0.4963}\\                
\bottomrule
\end{tabular}
\caption{Ablation Study on \textit{local hospital} dataset}
\label{table4}
\vspace{-10pt}
\end{table*}
 
 \subsection{Data Processing and Implementation}
 
 To evaluate our model on multiple time horizons, we selected different time lengths on \textit{local hospital} and MIMIC III datasets as survival time prediction targets.
 In the dataset from \textit{local hospital}, mortality in the first 90 days is most clearly related to immediate postoperative care. Mortality is divided into four groups: 0-6 days, 7-18 days, 19-30 days, and 31-90 days. For ICU datasets, death in the following day highlights potential immediately needs or patient deterioration. We therefore divide the MIMIC III mortality events into 0-6 hours, 6-12 hours and 12-24 hours after ICU admission as time-associated tasks. For ICU LOS, our applied question revolves around resource utilization and potentially avoidable ICU admissions, so we divide discharge times into 0-1 days, 1-2 days, and 2-4 days with 0-1 days as the main target. We include patients who die before discharge in the ``long stay'' category. 
For the purpose of selecting reference tasks, we use correlation analysis to identify several outcomes associated with the target.
 
For each above datasets, models are trained using 70\% vs. 30\% train-test split.
Networks used in our evaluation are four-layer fully-connected neural networks. The experiments are repeated five times with this division ratio, and the average performance is reported. 5-fold cross validation is used to fine-tune the network configuration and hyper-parameters including augmentation ratio, weight ratio ($w_k$ of time-associated tasks over $w_k$ of reference tasks) and $\alpha_k$. Adam~\cite{kingma2014adam} is the optimizer for inner and outer level updates. For three sets of experiments: survival time on \textit{local hospital}, survival time on MIMIC III, and ICU LOS, we choose 1e-3, 3e-4 and 3e-4 as the corresponding step size; 1.5, 1.2, 1.4 as the augmentation ratio; 1.3, 1.4, 1.4 as the weight ratio.

\subsection{Performance Evaluation}
 
We present AUROC and average recall as evaluation metrics. Baselines include a Deep Neural network (DNN), DNN-based semi-parametric survival analysis model (Survival), Multi-task Learning network (MTL), Pre-trained Model (Pre-train), Model agnostic meta learning (MAML), and a prototypical network (ProtoNet). The description of baselines is provided in the appendix.

The performance comparison is shown in Table \ref{table1}, \ref{table2}, and \ref{table3}, where TAML consistently outperforms all the other models. The deep neural network and survival analysis, which are trained separately for each time slot, has no advantages compared to other baselines. We observe that meta learning models (especially ProtoNet) and multi-task learning models always outperform other baselines. The reason why multi-task learning and prototypical networks can achieve the advantage is thanks to the division of tasks and the shared information between tasks. Compared to other baselines that make separate predictions for each time period, these two algorithms both jointly learn. 

The improvement of TAML in the second and third experiments is generally larger than that in the first experiment. As the time slot division in these two experiments contributes to much closer relationships between time periods, tasks in other time periods can provide more information under the TAML framework. We also observe that, compared to DNN, some predictions targeting the middle time period can be improved more than the two ends of the time period, for instance, the case of 19–30 days in the first experiment.

\subsection{Ablation Study}
In TAML, we propose several strategies to improve the MAML model for predicting time-associated tasks. The following ablation study is provided to evaluate the contribution of different components. The strategies include a weighted gradient update to emphasize the difference between time-associated tasks and reference tasks in the meta-train phase and the TISS label setting applied in the training process.
We aim to test whether both strategies are beneficial by removing each and redoing the experiments. We also explore the  sensitivity of the model to task selection. Results are shown in Table \ref{table4}, \ref{table5} and \ref{table6}. 
Since the results are consistent across the three experiments, the ablation study of the second and third experiments is in the appendix.

In \textbf{TAML w/o TISS} scenario, we omit TISS in both meta-train and meta-test phases, and only keep the weight assignment to time-associated tasks and reference tasks. In \textbf{TAML w/o weight} scenario, weights assigned to different tasks are the same. In \textbf{TAML w/o TISS train} scenario, TISS is omitted only in the meta-train phase. Besides, to assess the sensitivity of the model to task selection, the mutual information between tasks and the true target is calculated to filter out a few tasks that are least relevant to the current target. In \textbf{TAML w/o unrelated tasks} scenario, some tasks with low relevance are removed from training (See appendix for low relevance task list).

\begin{figure}[t]
	\centering
	\subfigure{
		\begin{minipage}[b]{0.228\textwidth}
			\includegraphics[width=1\textwidth]{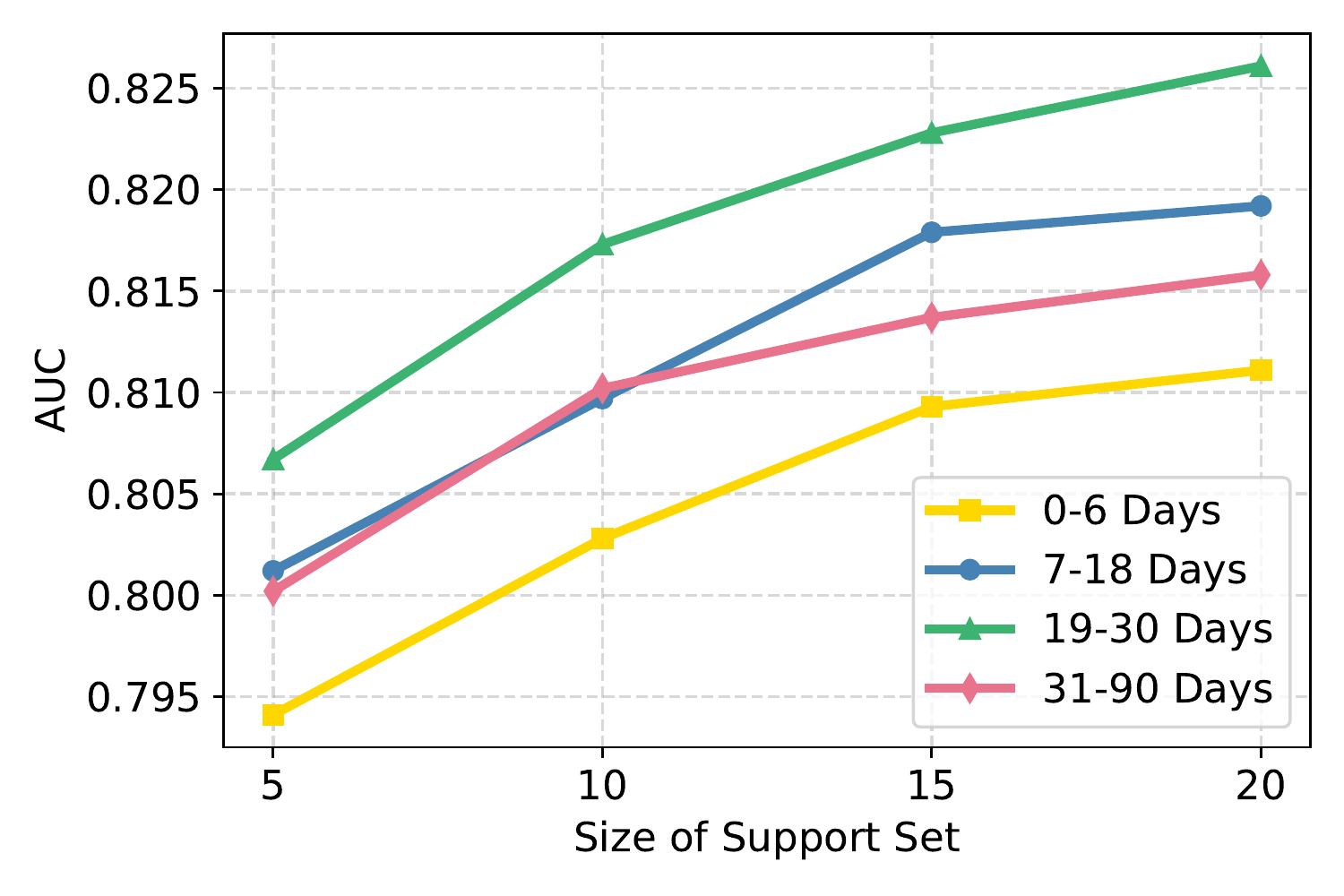} 
		\end{minipage}
	}
	\subfigure{
		\begin{minipage}[b]{0.228\textwidth}
			\includegraphics[width=1\textwidth]{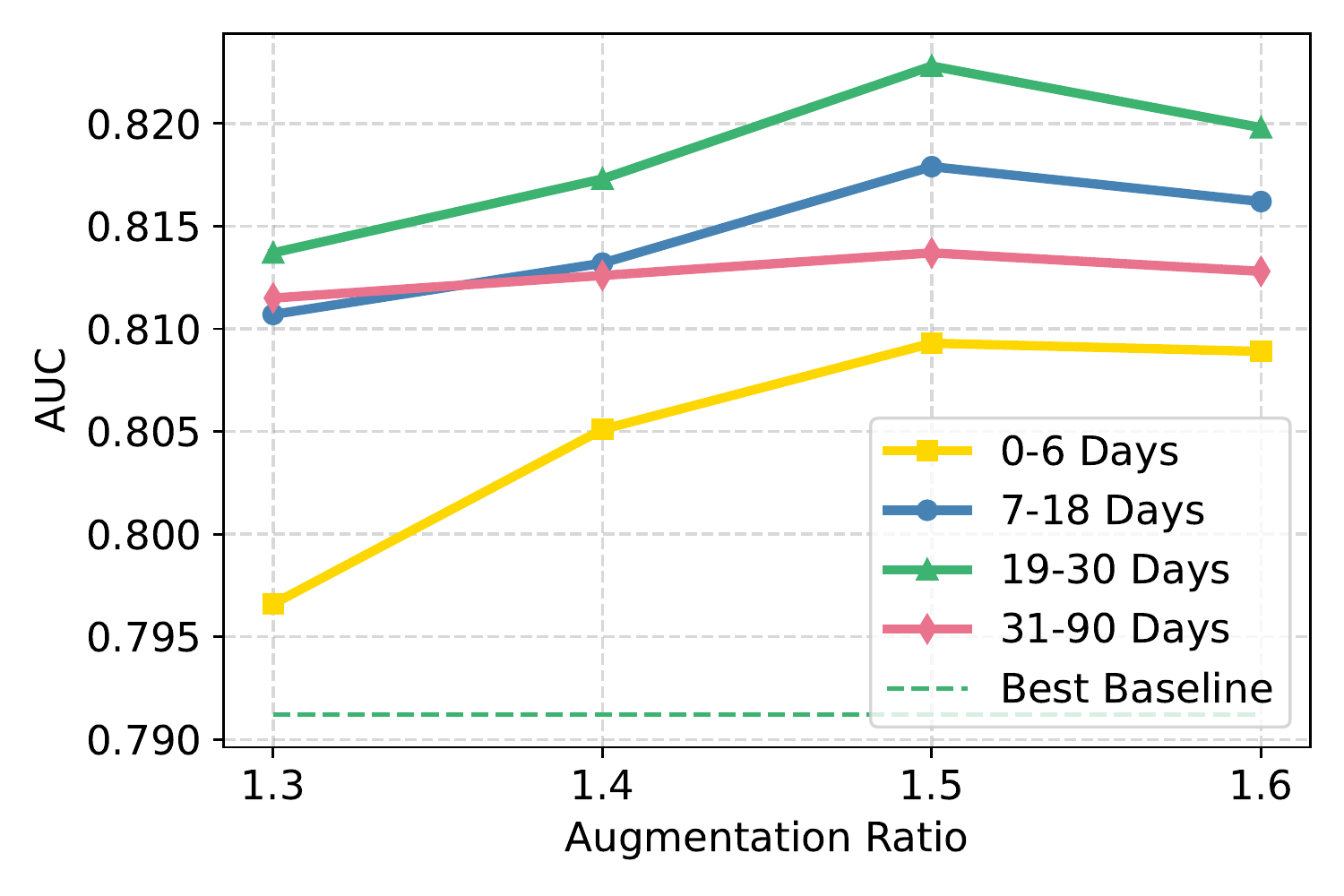} 
		\end{minipage}
	}
     \vspace{-1.5em} 
 
	\subfigure{
		\begin{minipage}[b]{0.228\textwidth}
			\includegraphics[width=1\textwidth]{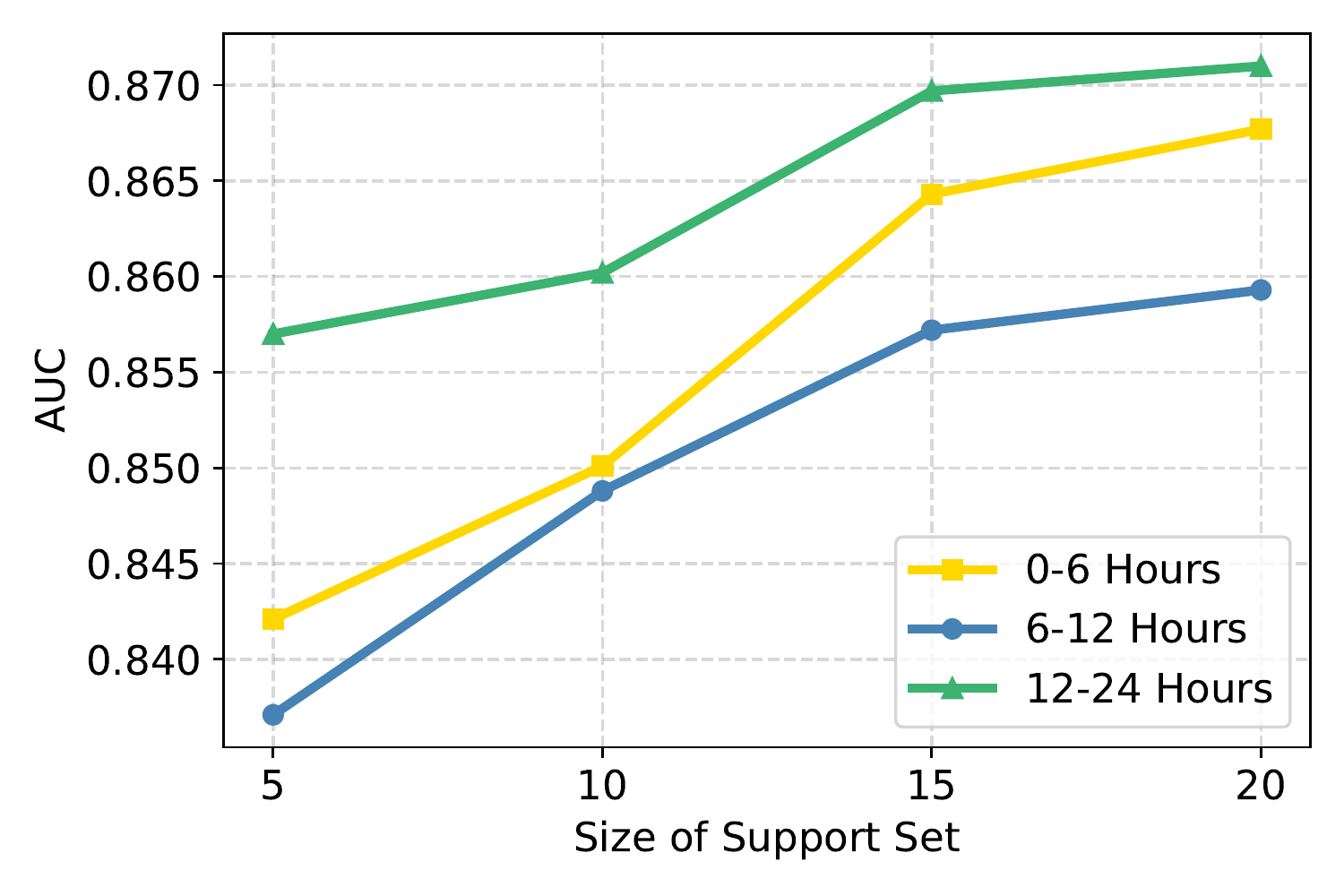} 
		\end{minipage}
	}
	\subfigure{
		\begin{minipage}[b]{0.228\textwidth}
			\includegraphics[width=1\textwidth]{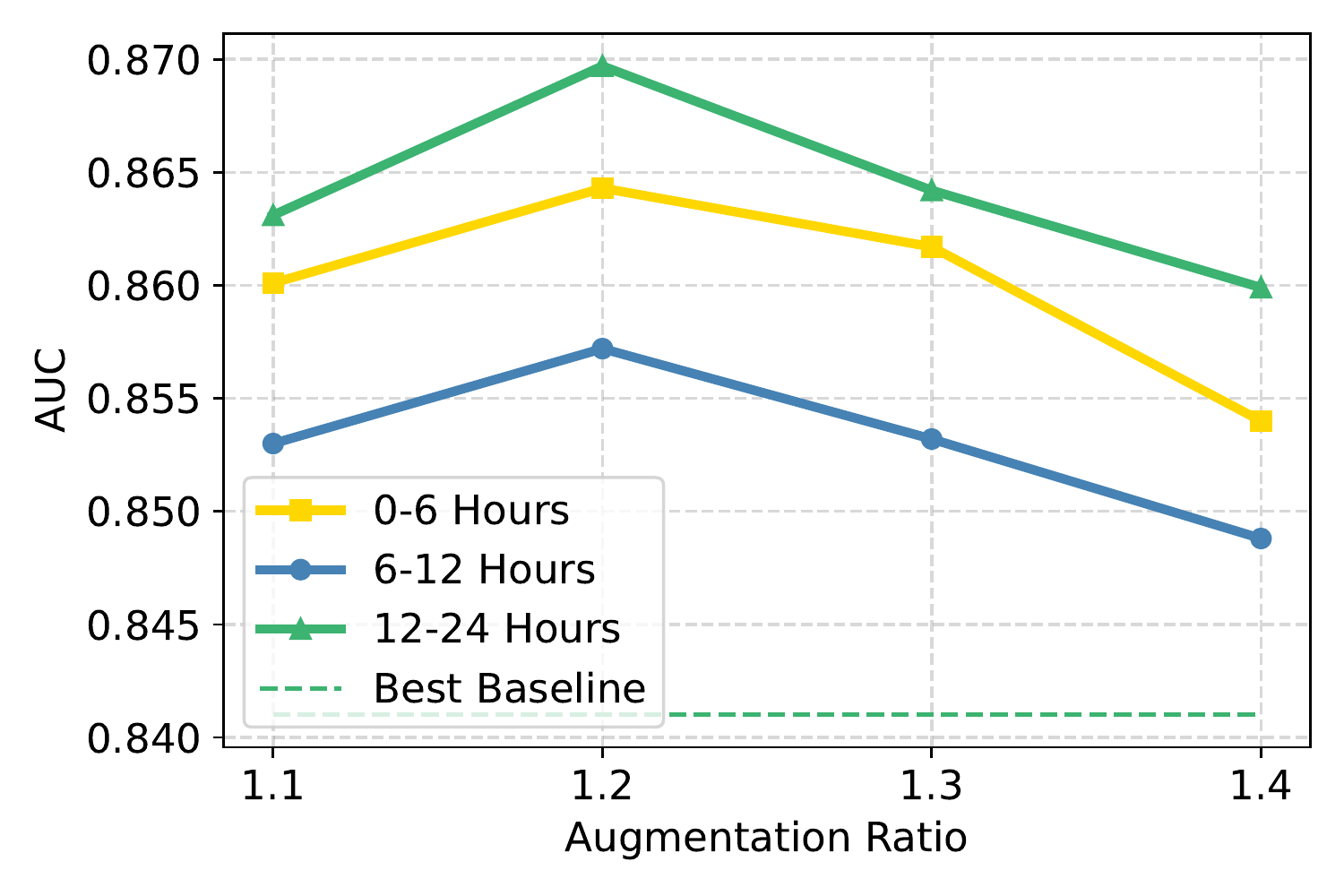} 
		\end{minipage}
	}
	\vspace{-1.5em} 
	
	\subfigure{
		\begin{minipage}[b]{0.228\textwidth}
			\includegraphics[width=1\textwidth]{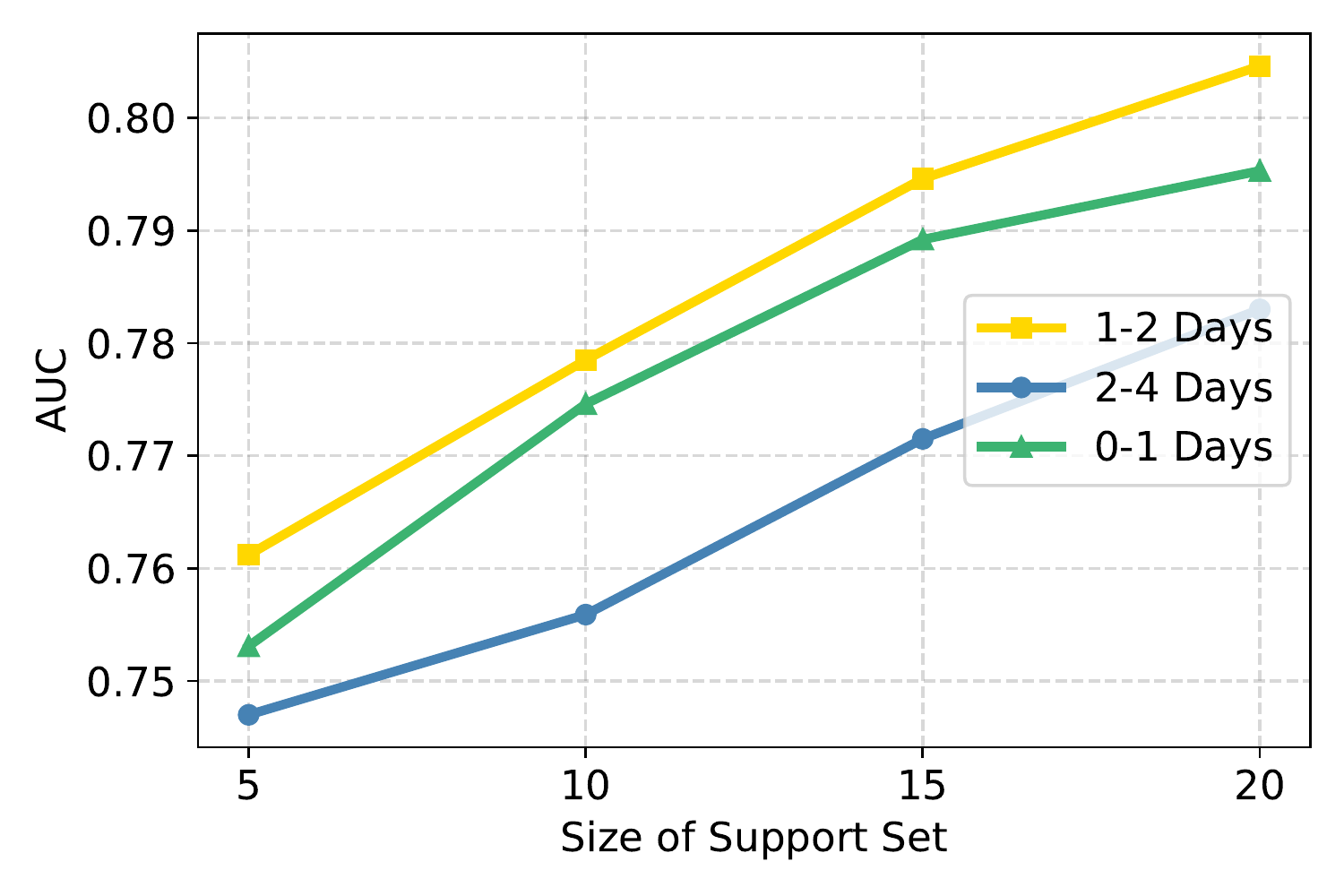} 
		\end{minipage}
	}
	\subfigure{
		\begin{minipage}[b]{0.228\textwidth}
			\includegraphics[width=1\textwidth]{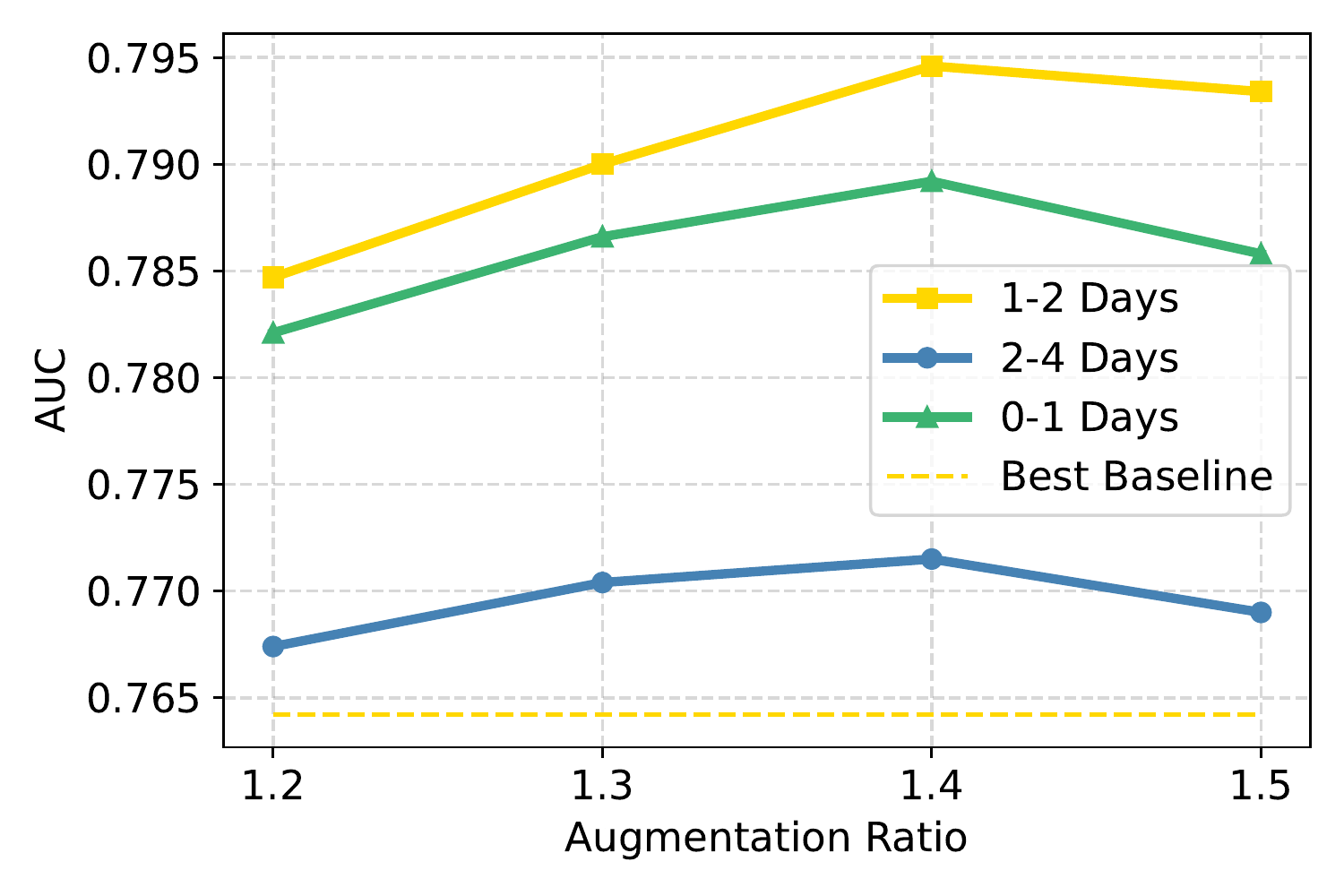} 
		\end{minipage}
	}
	\vspace{-1.5em} 
	\label{fig:grid_4figs_1cap_4subcap}
	\caption{Parameter Sensitivity Study. 
	Top: Survival time on \textit{local hospital}; Mid: Survival time on MIMIC III; Bottom: ICU LOS. First column: Size of Support Set; Second column: Situation Ratio. The dotted line is the best baseline of the same color target.}
\end{figure}

We can observe that each part of our framework has an enhancing effect on the results, but the enhancement of just assigning weights to different tasks is limited. The most powerful part of the TAML framework is the combination of temporal information sharing strategy in the meta-test phase and weight assignment among tasks. 
It gives an insight into the TAML framework: for targets of different time periods, the model can also have good performance if implementing the meta-test based on the same meta-train results. 

When some of the tasks with low relevance are removed, the change in the model performance is relatively small in all four time periods. This shows that although using some low-relevance tasks may induce noise in the model, it also ensures the model's generalization, so that model performance may not be hurt even if some low-relevance tasks are included in the selection of tasks.

\subsection{Sensitivity Analysis}

To evaluate the performance under different model settings and hyperparameters, we study the impact of the size of support set, augmentation ratio, train-test split ratio, and time window width and provide some rules for parameter selection. We show the results of the analysis on the first two parameters in Figure 3 and the rest of analysis in appendix.

\subsubsection{Size of Support Set:} The first column of Figure 3 shows that as the size of the support set increases from 5 to 20, the performance improves; however the smaller support set still has acceptable performance. The additional improvement from support tasks does not appear to saturate. We choose 15 as the size of support set in all experiments.

\subsubsection{Augmentation Ratio:} Observed from the second column, on the \textit{local hospital} dataset, the curve of augmentation ratio is nearly flat around 1.5, which indicates that paying more attention to ``real'' label tasks improves performance. However, on the MIMIC III dataset, the AUC drops when the value of the augmentation ratio is larger than 1.2, suggesting that the auxiliary label samples are helping. The reason might be that in the MIMIC III data set the time periods are contained in a fairly narrow window, thus more smoothing is helpful compared to the \textit{local hospital} dataset. Thus, a larger augmentation ratio can be chosen for the setup with weak time period correlation.

Additionally, we add a line of the best criterion in the form of a dashed line, which represents the best baseline result of the same color target. Although there are small fluctuations in AUC as the augmentation ratio changes, the advantage of TAML over the best baseline is still significant.

\begin{figure}
    \vspace{-10pt}
    \centering
    \includegraphics[width=0.47\textwidth]{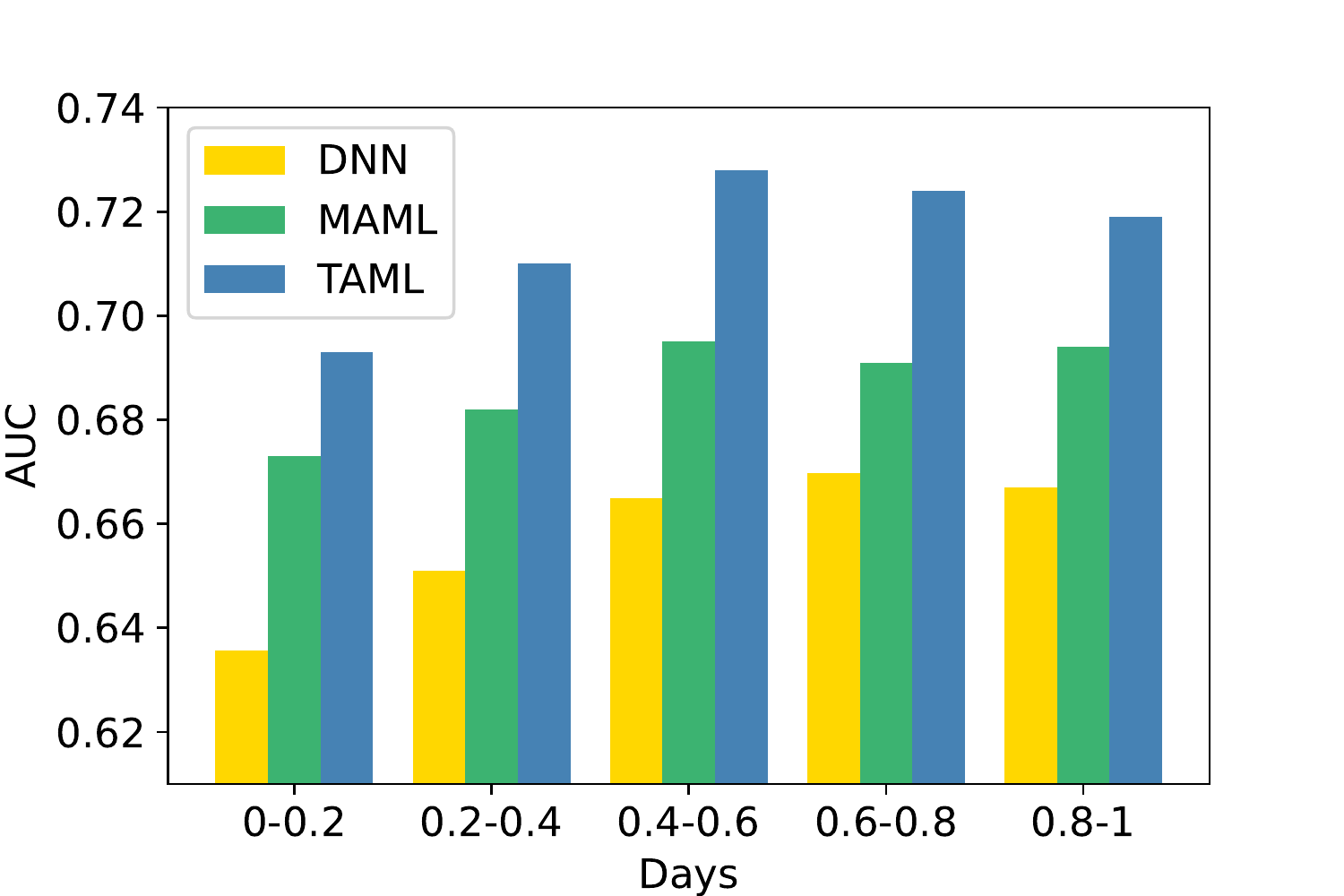}
    \caption{Short Time Slot experiments on ICU LOS}
    \label{fig:shortLOS}  
\end{figure}

\subsubsection{Target time window width:} We vary the width of the target window in ICU LOS from 0.2 to 1 day, showing the results in Figure~\ref{fig:shortLOS}. With small target windows, TAML outperforms other baselines, although there is a small decrease in performance. Compared to MAML, TAML has more improvement in terms of learning ability for the middle time widths, which benefit from the TISS.

\section{Conclusions}

In this paper, we focus on a specific few-shot problem in clinical predication: predicting future events in specific time-ranges (time-associated target prediction). We propose a time-associated meta learning (TAML) framework. We increase the choice of available tasks by transforming the regression problem into multiple binary classification problems, adding not only many closely related time-associated tasks, but also some time-independent reference tasks. To deal with the sparsity caused by task splitting, the temporal information shared strategy (TISS) is designed to augment positive labels and smooth the relationship between adjacent event categories during training.

We validate our model on public datasets and \textit{local hospital} datasets to predict two clinical events: survival time and ICU length of stay (LOS), on which our model shows strong performance over baselines. Additionally, TAML turns out to be insensitive to tuning parameters and unrelated tasks and can achieve excellent performance regardless of the length of the time window. Thus, the TAML framework can be applied to clinical time-associated target prediction, thereby providing a reference for decision-making. Besides, we develop an Python package MetaEHR for few-shot clinical prediction, which includes the implementation of TAML on time-associated prediction as well as other meta learning algorithms for time-independent prediction.

\appendix
\section{Appendix}
\subsection{Baseline Description}

\noindent\textbf{Simple DNN}
Traditional deep neural network is implemented as a baseline, where the patient EHR data is used as input, mortality or ICU LOS in a selected time period is considered as target. We train a separate DNN for each target and report the results in turn.

\noindent\textbf{MAML} 
The MAML framework is used as a baseline, where we treat time-associated tasks and reference tasks as indistinguishable. In the training process, different situations are given the same weight and TISS is also not used.

\noindent\textbf{Multi-task Learning}
A shared-layer multi-task learning network is considered as a baseline, where the prediction targets are mortality in different time periods.

\noindent\textbf{Survival Analysis}
DNN-based semiparametric survival analysis which predict the event time is taken to evaluate. For a comparable evaluation, we divide the predicted time into the same groups in TAML and
evaluate the performance in each group.

\noindent\textbf{Prototypical Neural Network}
Prototypical neural network is used as a baseline representing metric-based meta learning. The number of prototypes equals to the number of time periods in each dataset.

\noindent\textbf{Pretrained Model}
We also pre-train a model to compare the performance, where all the tasks involved in TAML are used in pre-training process. Then we fine-tune the model on the time-associated tasks.

\subsection{Ablation study on the second and third experiment}

The ablation study on the second and third experiments is provided in Table \ref{table5} and \ref{table6}.

\begin{table}[H]
\centering
\small
\setlength\tabcolsep{2pt}
\begin{tabular}{lcccccc}
\toprule
\multicolumn{1}{c}{\multirow{2}{*}{Model}} & \multicolumn{2}{c}{0-6 Hours}     & \multicolumn{2}{c}{6-12 Hours}    & \multicolumn{2}{c}{12-24 Hours}   \\\cmidrule(l){2-3}\cmidrule(l){4-5}\cmidrule(l){6-7}
\multicolumn{1}{c}{}                       & AUROC           & Recall          & AUROC           & Recall          & AUROC           & Recall          \\
\midrule
MAML                                       & 0.8297          & 0.4598          & 0.8216          & 0.4571          & 0.8304          & 0.4694          \\
w/o TISS                                       & 0.8374          & 0.4733          & 0.8299          & 0.4698          & 0.8465          & 0.4802          \\
w/o weight                                       & 0.8423          & 0.4701          & 0.8419          & 0.4705          & 0.8521          & 0.4749          \\
w/o train
& 0.8592 & 0.4893 & 0.8476 & 0.4801 & 0.8616 & 0.4971\\
w/o unrelated &
0.8671 & 0.4867 & 0.8532 & 0.4847 & 0.8685 & 0.5028\\
\textbf{TAML(ours)}                        & \textbf{0.8643} & \textbf{0.4891} & \textbf{0.8572} & \textbf{0.4870} & \textbf{0.8697} & \textbf{0.4958}\\
\bottomrule
\end{tabular}
\caption{Ablation Study of Survival Time on MIMIC III dataset}
\label{table5}
\end{table}
\begin{table}[H]
\centering
\small
\setlength\tabcolsep{2pt}
\begin{tabular}{lcccccc}
\toprule
\multirow{2}{*}{Model} & \multicolumn{2}{c}{0-1 Days}      & \multicolumn{2}{c}{1-2 Days}      & \multicolumn{2}{c}{2-4 Days}      \\ \cmidrule(l){2-3}\cmidrule(l){4-5}\cmidrule(l){6-7}
                       & AUROC           & Recall          & AUROC           & Recall          & AUROC           & Recall          \\ \midrule
MAML                   & 0.7458          & 0.4209          & 0.7642          & 0.4278          & 0.7486          & 0.4236          \\
w/o TISS        & 0.7625          & 0.4353          & 0.7781          & 0.4489          & 0.7628          & 0.4337          \\
w/o weight        & 0.7639          & 0.4317          & 0.7725          & 0.4458          & 0.7623          & 0.4346          \\
w/o train  & 0.7801          & 0.4412          & 0.7876          & 0.4548          & 0.7684          & 0.4392          \\ 
w/o unrelated & 0.7875 & 0.4495 & 0.7993 & 0.4662 & 0.7723 & 0.4422\\
\textbf{TAML(Ours)}    & \textbf{0.7892} & \textbf{0.4418} & \textbf{0.7946} & \textbf{0.4617} & \textbf{0.7715} & \textbf{0.4436} \\ \bottomrule
\end{tabular}
\caption{Ablation Study of ICU LOS}
\label{table6}
\end{table}

\subsection{Parameter Sensitivity: Train and test split}
The parameter sensitivity of the parameter: train and test split ratio is provided in Figure 5.

\begin{figure}[h]
	\centering
    	\subfigure{
    		\begin{minipage}[b]{0.228\textwidth}
   		 	\includegraphics[width=1\textwidth]{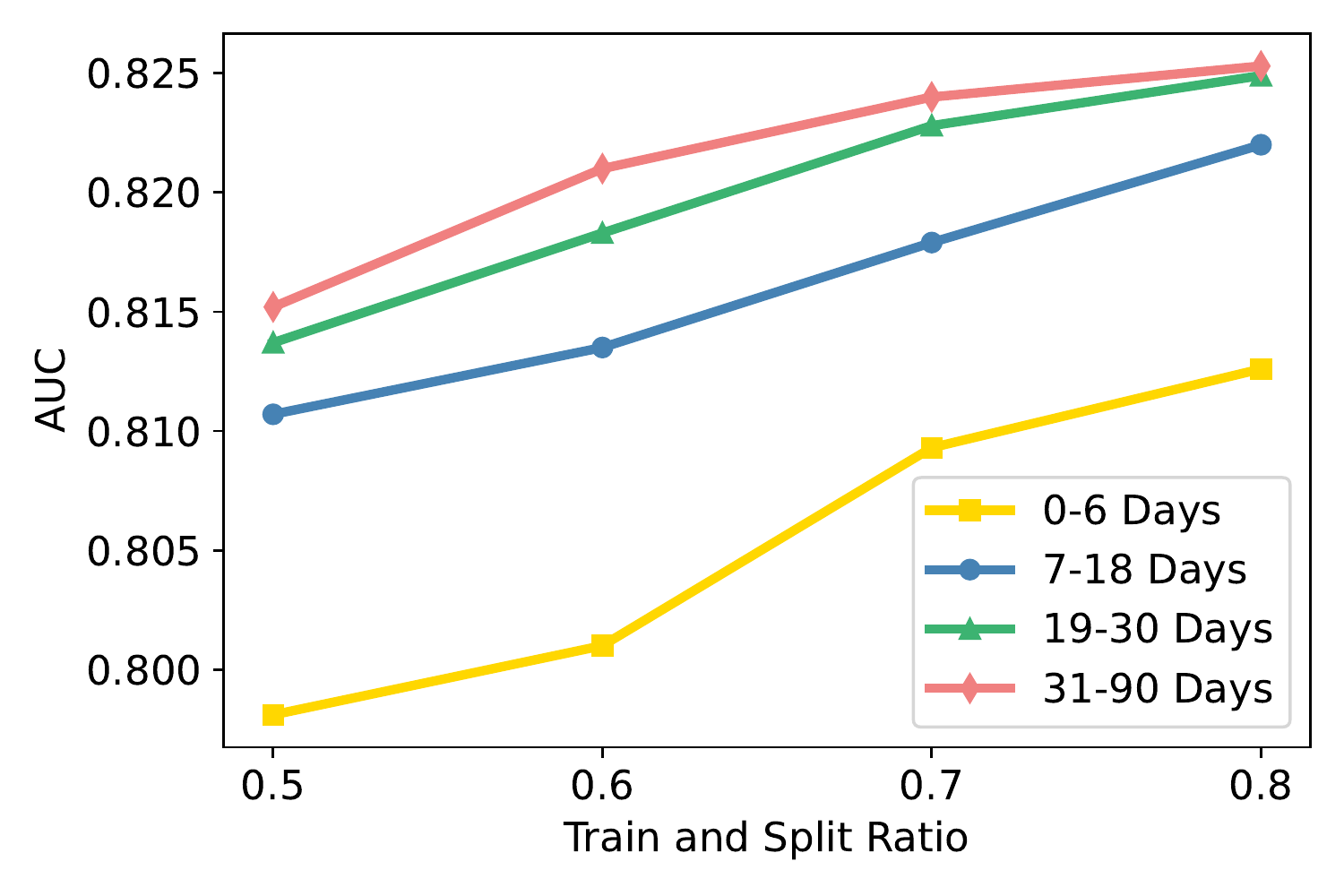}
    		\end{minipage}
    	}
    	\subfigure{
    		\begin{minipage}[b]{0.228\textwidth}
   		 	\includegraphics[width=1\textwidth]{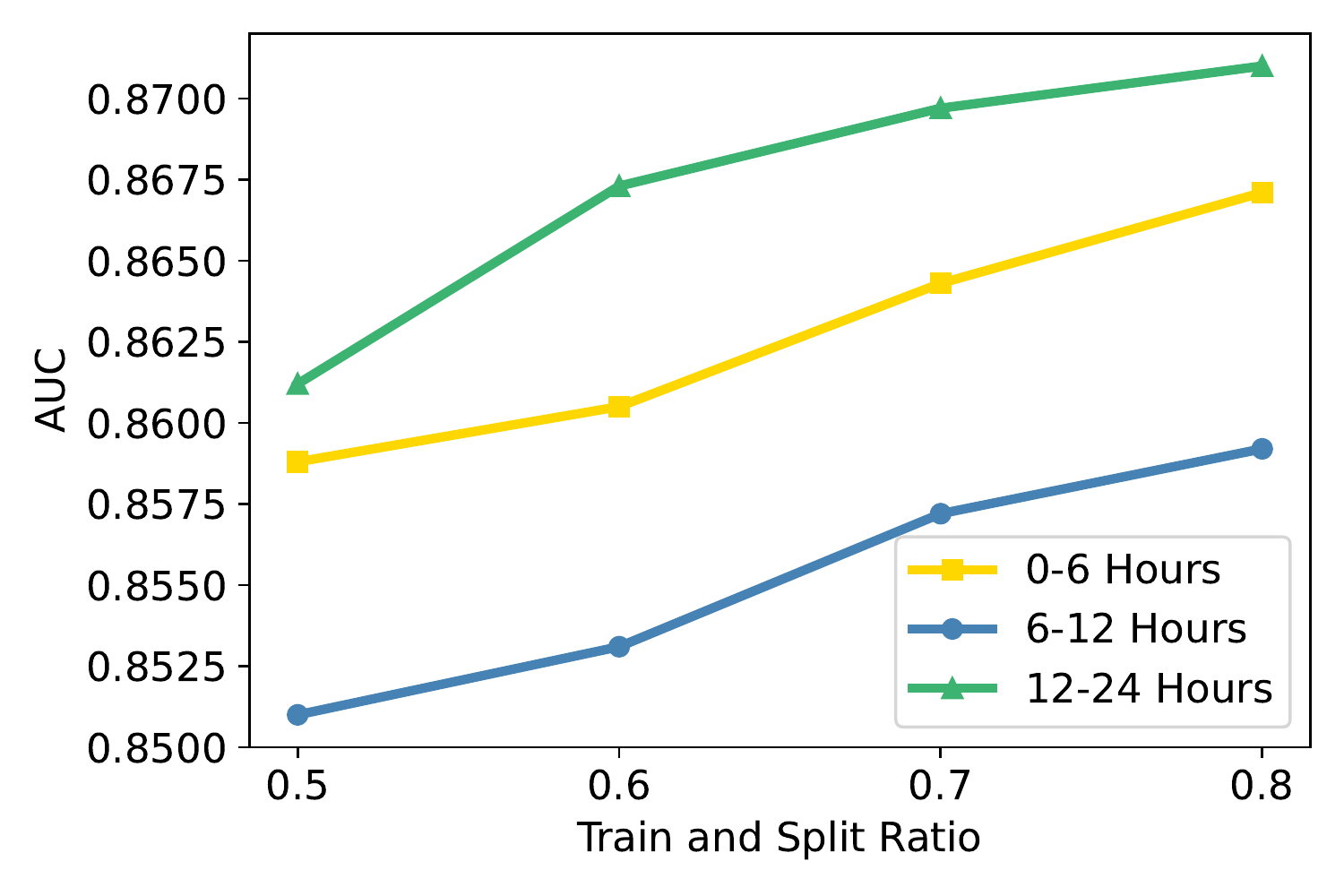}
    		\end{minipage}
    	}
	
	\label{fig:sa2}
	\caption{Parameter Sensitivity Study. Left: Survival time on \textit{local hospital}, Right: Survival time on MIMIC III
	}
\end{figure}

\subsection{Reference Tasks}

 In \textbf{survival time experiment on \textit{local hospital}}, we use 13 reference tasks including: ARF, Heart Attack, Cardiac Arrest, CHF, Stroke, LegBloodClot, LungBloodClot, Respiratory Arrest, Pneumonia, GIBleed, Abnormal Heart Rythmn, VTE, disposition, where we select four low-relevance tasks by calculating mutual information. These four tasks are LegBloodClot, LungBloodClot, Respiratory Arrest, Stroke.

In \textbf{survival time experiment on MIMIC III} dataset, we use 12 reference tasks including: Speticemia, Cardiac dysrhythmias, AKI, heart failure, peripheral vascular, hypertension, diabetes, liver disease, myocardial infarction (MI), coronary artery disease (CAD), cirrhosis, and jaundice, where we select four low-relevance tasks. These four tasks are peripheral vascular, hypertension, diabetes, and jaundice.

 In \textbf{ICU LOS} experiment, we use 13 reference tasks including:
Arrhythmia, Stroke, Hyperglycemia, HepaticDisease, AnemiaBleed, Coagulopathy, Thrombocytopenia, Pneumonia, UTI, BloodInfection, Sepsis, AlteredMentalStatus, RespFailure, where we select four low-relevance tasks. These four low relevance tasks are BloodInfection, Thrombocytopenia, Coagulopathy, and HepaticDisease.

\bibliographystyle{named}
\bibliography{ijcai22}

\end{document}